\documentclass[runningheads]{llncs}
\usepackage{graphicx}
\usepackage{algorithm}
\usepackage{algorithmic}
\usepackage{graphicx}
\usepackage[caption=false]{subfig}
\usepackage{amsmath}
\usepackage{mathtools}
\DeclarePairedDelimiter{\ceil}{\lceil}{\rceil}
\begin{document}
\title{A State Aggregation Approach for Solving Knapsack Problem with Deep Reinforcement Learning}
%
%
\author{Anonymized for review}
\author{Reza Refaei Afshar\inst{1}\orcidID{0000-0003-1558-8380} \and
Yingqian Zhang \inst{1}\orcidID{0000-0002-5073-0787} \and
Murat Firat \inst{1}\orcidID{0000-0001-8483-3533} \and
Uzay Kaymak \inst{1}\orcidID{0000-0002-4500-9098}}
%
%
\institute{Eindhoven University of Technology, Eindhoven, Netherlands}
%
\maketitle              
\begin{abstract}
This paper proposes a Deep Reinforcement Learning (DRL) approach for 
solving knapsack problem. 
The proposed method consists of a state aggregation 
step based on tabular reinforcement learning to extract features and construct states. 
The state aggregation policy is applied to each problem instance of the knapsack problem, which 
is used with Advantage Actor Critic (A2C) algorithm to train a policy through which the items are sequentially selected at each time step. The method is a constructive solution approach and the process of selecting items is repeated until the final solution is obtained. The experiments show that our approach provides close to optimal solutions for all tested instances, outperforms the greedy algorithm, and is able to handle larger instances and more flexible than an existing DRL approach. In addition, the results demonstrate that the proposed model with the state aggregation strategy not only gives better solutions but also learns in less timesteps, than the one without state aggregation.  

\end{abstract}
\section{Introduction}
\label{sec:introduction}

Heuristic algorithms for solving Combinatorial Optimization Problems (COPs) achieve acceptable solutions in a polynomial time. These algorithms relies on handcrafted heuristics that conduct the process of finding the solutions. Although these heuristics work well in many COPs, they mostly rely on the nature of problems and they need to be modified for different class of problems. In this paper, 
we aim to learn and improve the handcrafted heuristics to improve the quality of the solutions. We study \textit{knapsack problem (KP)}, which is one of the well-known benchmark problems in COPs. KP is defined as a set of items, each with a value and a weight. The objective is to select a subset of items with maximum total value to fill a knapsack such that the cumulative items weight does not exceed its capacity. This problem has many applications such as cargo loading, cutting stock and capital budgeting \cite{wilbaut2008survey}.

Recently, there is a great progress in the Artificial Intelligence (AI) community in developing machine learning (ML) methods to solve COPs \cite{bengio2018machine}, where a popular ML based method is \textit{Deep Reinforcement Learning (DRL)}. DRL is the integration of \textit{Reinforcement Learning (RL)} and \textit{Deep Neural Networks (DNN)} \cite{arulkumaran2017brief,lecun1998gradient,huang2019review}. 
Several DRL based approaches have been proposed to solve the Traveling
Salesman Problem (TSP), e.g. \cite{bello2017neural,joshi2019efficient,kool2018attention}, where a discrete representation of TSP is used as states and the solution is a sequence of the inputs. These approaches work well for the TSP problem, however, in the Knapsack problem, the values and weights of items are continuous, which entails an extremely large state space when the number of items increases. Hence, the existing DRL based approaches for solving the problems with discrete nature such as TSP might not work well for KP, as shown in 
\cite{bello2017neural}. The authors of \cite{bello2017neural} solve a Knapsack problem using the policy gradient algorithm with pointer networks. Although they show optimal solutions can be obtained for instances up to 200 items, the following limitations are identified: (1) intractability to large instances: the state space grows rapidly with increasing number of items, and (2) generality to other sizes of instances: the trained model is applicable for solving the problems that have the exactly same knapsack capacity and same number of items. In this paper, we introduce a DRL approach with state aggregation that boosts the capability of the typical greedy algorithm and improves the heuristic to overcome these two limitations. 


In our approach, 
propose a state aggregation method to discretize the feature values of items. 
We construct a feature table by assigning a row for each problem instance and a column for each item's  information. A tabular reinforcement learning is used to learn the best operation strategy for each item. 
This resulting discretization of features not only provides a discrete representation of the problem instances, but also reduces the state space by reducing the number of unique values.
However, the state space is still large despite state aggregation.   
Therefore we exploit DRL as a powerful function approximation approach. We use \textit{Advantage Actor Critic (A2C)} algorithm to learn the policy of selecting items. A2C makes use of two DNNs for learning policy and value functions \cite{mnih2016asynchronous}. The policy DNN has $N$ outputs which is equal to the number of items. By following a greedy or softmax algorithm on the output of the policy DNN, a sequence of items are selected until the knapsack is full.

The experimental results show that the proposed approach finds optimal solutions with two decimal places for the problem instances of same size used in \cite{bello2017neural}. Moreover, we show the method obtains close to optimal solutions for three different types of instances with at most 50, 300 and 500 items. 
We also demonstrate that the proposed DRL with state aggregation performs better than the DRL without aggregation in terms of both learning rate and the solution quality. We summarize our contributions as follows.
\begin{itemize}
\item We develop a state aggregation strategy to derive state embedding that reduces the state space size. We show this general strategy effectively speed up learning on solving KP.
\item Our DRL-based approach to solve KP improves the heuristic greedy algorithm for 0-1 KP and shows better performance than the existing DRL approaches. The developed method can be trained once for $N$ items and it can be used for any KP instances with size up to $N$.
\end{itemize}


\section{Related Work}
\label{sec:related}

It has been proven by reduction that most of COPs are NP-Hard problems. Their optimal solutions can not be found in polynomial time and exact algorithms take exponential time to find optimal solutions \cite{karp1972reducibility,cook2006p}. Knapsack Problem (KP) has gained a remarkable attention in the literature. Despite the fact that the fractional KP is optimally solvable by the heuristic greedy algorithm, the 0-1 knapsack problem is NP-Hard \cite{cormen2009introduction}, and a large variety of KPs remain hard to solve \cite{pisinger2005hard}. Moreover, it has been shown by empirical evidence that solving instances near the phase transition are challenging for humans \cite{yadav2018phase}. The phase transition emerges around critical values of items and capacity so that the probability of having a solution for an instance change from zero to one. Many algorithms, ranging from dynamic programming algorithms (e.g. \cite{dasgupta2008algorithms}) to meta-heuristics (e.g. \cite{feng2018solving}) have been proposed to solve KP.   

Cleverly searching and branch and bound methods can prune the search tree and reduce the time complexity of COPs \cite{woeginger2003exact}. However, these methods are still prohibitive for large instances. Polynomial time approximation schemes and integer linear programming (ILP) based approaches are the other helpful methods \cite{vazirani2013approximation,du2013handbook}. Although the approximation algorithms might be performed in reasonable time, they rely on handcraft heuristics and the methods need to be revised when the problem settings change. Furthermore, they suffer from weak optimality for some problems. In order to cope with this limitations, Machine Learning (ML) based and data driven methods are developed.

In recent years, it has been shown that DRL can be used for learning good heuristics for solving COPs. In \cite{vinyals2015pointer}, the Pointer Network architecture is introduced where the output layer of the deep neural network is a function of the input. In \cite{bello2017neural}, the pointer network is used with RL to solve the Traveling Salesman Problem (TSP). They use policy gradient and a variant of Asynchronous Advantage Actor-Critic (A3C) algorithm of \cite{mnih2016asynchronous} to train a DNN, and show close to optimal solutions are found for up to 100 cities. In \cite{khalil2017learning} another neural network framework is introduced for graph-based COPs, where \textit{structure2vec} \cite{dai2016discriminative} is used to derive an embedding for the vertices of the graph. The structure2vec computes a p-dimensional feature embedding for each node and a parametric $Q$ function is trained using Q learning algorithm. In \cite{kool2018attention}, the pointer network is incorporated with attention layers. With the REINFORCE algorithm, they obtained close to optimal solutions for the TSP instances of up to 100 nodes.

Most of ML-based research on solving COPs focuses on TSP. COPs like TSP and Vehicle Routing Problem that have gained high attentions in past few years, require a sequence of the input as the solution and sequence-to-sequence neural architectures might be proper approaches for solving them \cite{sutskever2014sequence}. 
However, the solutions of COPs like KP and Weighted Vertex Cover are a subset of the input. This issue makes the original sequence-to-sequence approaches inapplicable for solving KP. Recently, a pointer network deep learning approach is presented for solving 0-1 KP \cite{gu2018pointer}. This method is based on supervise learning and optimal solutions which is not available in most of the cases. In this paper we propose a DRL framework for subset selection problems.

\section{Problem Definition and Modeling}
\label{sec:prob-definition}
We consider the following \textit{0-1 Knapsack Problem} instance $P$: We are given a set $\mathcal{I_P}$ containing $n_P$ items and a knapsack of capacity $W_P$. Each item $i$ has value $v_i$ and weight $w_i$. The goal is to fill the knapsack with a selected subset of items such that the total weight of the selected items does not exceed $W_P$ and the total value is maximized. Since $P$ is a 0-1 KP, selecting a fraction of an item is not possible.

Our method for solving this variant of KP is based on deep reinforcement learning. We assume that the number of items is variable and a constructive solution can solve the problem only by considering the capacity constraint. Therefore, the process of selecting a subset of items $\mathcal{I'_P}\subseteq{\mathcal{I_P}}$ is modeled as a sequential decision process. The policy DNN is trained with A2C introduced in \cite{mnih2016asynchronous} on a set of problem instances with at most $N$ items. The information of each problem instance consists of $|\mathcal{I_P}|=n_P\leq{N}$ items with value $v_i$ and weight $w_i$ for each $i\in{\mathcal{I_P}}$ and together with $W_P$, they are the inputs of DNN. The DNN has $N$ outputs that each being associated with a value of selecting a specific item $i\in\mathcal{I_P}$. The policy is to select an item in each step. After selecting item $i$, it is removed from the original problem instance $P$ and a new problem instance $P'$ with a reduced item set $\mathcal{I_{P'}}=\mathcal{I_{P}}\setminus{\{i\}}$ and capacity $W_{P'}=W_P-w_i$ is generated. For the cases where $i$ cannot be added to the knapsack because of the capacity constraint, the new instance $P'$ is generated by simply removing $i$ from the item set, without altering $W_P$. In this way, when the policies are trained with problem instances of at most $N$ items, the policies can be used to find solutions for new instances as long as their item sizes are no greater than $N$.

Such KP problems can be found in different applications. For example, an online ad publisher faces with a set of advertisements. Assuming a fixed upper bound for the number of ads, the problem is to select a subset of them to show to the users. In this example, the values are relevance scores and the weights are the size of ads banners. The goal is to fill a slot of a certain size with the ads.


\begin{figure}
\includegraphics[width=\linewidth]{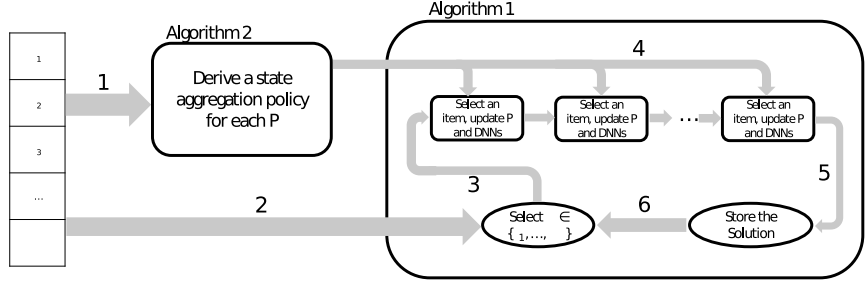}
\caption{The overview of the KP solver method. 1) A set of problem instances are used for deriving an aggregation policy for item information. 2) The same set of problems are used in the second step which is DRL. 3) A problem instance is selected for training. 4) Items are selected sequentially until finding a solution. At each step the updated $P$ is aggregated to find the state. The parameters of value and policy DNNs are updated using A2C. 5) The best solution is stored. 6) Another problem instance is selected for training. The process continues for a certain number of timesteps.}
\label{fig:knapsack-solver-overview}
\end{figure}


\section{DRL-based KP Solver}
\label{sec:method}

Figure \ref{fig:knapsack-solver-overview} shows the overview of our proposed method. It consists of two components. The first component includes a formulation of KP to MDP, which is solved using a DRL approach (Algorithm \ref{alg:knapsack-solver}).  The second component is a state aggregation method (Algorithm \ref{alg:aggregation}), which learns a aggregation policy to discretize states that are serves as inputs to DRL. We first discuss how to formulate the KP problem as MDP. 


\subsection{Deep Reinforcement Learning method}
\label{sec:drl}
In order to solve the 0-1 KP, DRL is used to derive a policy through that the items are sequentially added to the solution. We define the states, actions and rewards of DRL modeling of KPs for an instance $P'$ which is a representation of $P$ after selecting some items, as follows.

\paragraph{States $s(P)$:} A complete set of information of instance $P'$ containing $n_{P'}$, $v_i$ and $w_i$ for $n_{P'}$ items, capacity $W_{P'}$, the total value of the items ($Sv=\sum_{i\in \mathcal{I_{P'}}}{v_i}$), and the total weight of the items ($Sw=\sum_{i\in \mathcal{I_{P'}}}{w_i}$) makes a feature vector of $2n_{P'}+4$ features. Since $n_{P'}\leq{N}$ for all $P'$, the feature vector of the instances that have $n_{P'}<N$ items consists of $2N+4$ features such that the first $2n_{P'}+4$ ones carry the information of the problem instance and the remaining ones are zero. Section \ref{sec:rl-aggregator} will reduce this feature vector by a state aggregation strategy.
	
\paragraph{Actions:} There are $N$ actions $A_1,A_2, ..., A_N$, each corresponding to select one item. At each decision moment, a state is fed to the policy DNN and an action is selected according to the output of the DNN.

\paragraph{Reward Function:} The reward function is defined based on three criteria. First, if item $i$ can be added to the  knapsack without exceeding the capacity limit, $v_i$ is used as a positive reward. Second, if $w_i$ is greater than $W_{P'}$, i.e. $i$ cannot be added to the knapsack, $-w_i$ is set as a negative reward. Third, for each instance $P'$ where $n_{P'}<N$, the first $n_{P'}$ outputs of DNN correspond to the items of $P'$ and the next $N-n_{P'}$ outputs are undefined actions because the corresponding items do not exist. Therefore, a large penalty i.e. $-W_{P'}$ is used for the reward of choosing undefined actions. We separate the reward of undefined action and heavy items because an action with $i>n$ is always undefined, however items with $w_i>W_P$ could be added to the knapsack if they were selected in earlier steps. Therefore, their penalty is lower. Equation (\ref{eq:reward-DNN}) shows the reward of state $s(P')$ and action $A_i$.
\begin{equation}
\label{eq:reward-DNN}
    r(s(P'),A_i)=
\begin{cases}
	-W_{P'} & \mathrm{if}\;\;i>n_{P'} \\
       v_i & \mathrm{if} \;\; {w_i}\leq W_{P'}\\
       -w_i & \mathrm{if} \;\; w_i>W_{P'}
              \end{cases}
\end{equation}
 
 Employing these definitions of states, actions and rewards, the A2C algorithm is used for training policy and value DNNs \cite{mnih2016asynchronous}, where two DNNs are used for policy ($\pi$) and value ($V$) functions. The advantage value is obtained by subtracting state values (V) from state action values (Q) which is defined by $r+\gamma{V(s_{t+1})}$. This value is used in gradient function to update the parameters of the DNNs using Equations (\ref{eq:gradient-policy}) and (\ref{eq:gradient-value}) \cite{mnih2016asynchronous,stable-baselines}.

\begin{equation}
\label{eq:gradient-policy}
\theta^{t+1} \leftarrow \theta^t + \nabla_{\theta^t}\log\pi(A_i|s(P),\theta^t)
[r_t + \gamma{V(s(P'),\theta_v^t)}-V(s(P),\theta_v^t)]
\end{equation}

\begin{equation}
\label{eq:gradient-value}
\theta_v^{t+1} \leftarrow \theta_v^t + \frac{\partial(r_t + \gamma{V(s(P'),\theta_v^t)}-V(s(P),\theta_v^t))^2}{\partial{\theta_v^t}}
\end{equation}

where, $\theta^{t}$ and $\theta_v^{t}$ are the parameters of policy and value DNNs in decision moment $t$ respectively. The corresponding state of a problem instance $P$ is fed to the policy DNN and the items can be selected by following a policy according to the output of the policy DNN. Upon selecting an item, $P'$ is obtained from $P$ and it is again fed to the policy DNN to select the next item. This process is continued until filling the knapsack or exceeding the weight constraint. Algorithm \ref{alg:knapsack-solver} shows the DRL-based knapsack solver method.


\begin{algorithm}[t]
\caption{DRL-based Knapsack Solver}
\label{alg:knapsack-solver}
\textbf{Input}: $M$ Problem Instances each having at most $N$ items\\
\textbf{Output}: Values of solutions of the $M$ instances \\
\begin{algorithmic}[1] 
\STATE Initialize a policy DNN with $2N+4$ inputs, $N$ outputs and parameters $\theta$ as policy $\pi{(A_i|s,\theta)}$
\STATE Initialize a value DNN with parameters $\theta_v$ as $V(s,\theta_v)$
\STATE $t_{max}=3N\times{10^4}$, $t=0$
\STATE Initialize $Val$: a list of length $M$, all $0$
\WHILE{$t<t_{max}$}
	\STATE Select a problem instance $P$ with capacity $W_P$.
	\STATE $ow=0$ \COMMENT{Total weight of selected items}
	\STATE $ov=0$ \COMMENT{Total values of selected items}
	\STATE $P'\leftarrow{P}$, $n_{P'}\leftarrow{n_P}$, $W_{P'}\leftarrow{W_P}$
	\WHILE{$ow<W_{P'}$ and $n_{P'}>0$}
		\STATE Find $s(P')$ using state aggregation strategy (Eqn. (\ref{eq:states})) \label{line:s(p)}
		\STATE Perform action $i$ according to policy $\pi{(A_i|s(P'),\theta^t)}$ and observe $r(s(P'),A_i)$
		\IF {$i\leq{n_{P'}}$ and $w_i+ow\leq{W_{P'}}$}
			\STATE $ow\leftarrow{ow+w_i}$ 
			\STATE $ov\leftarrow{ov+v_i}$
			\STATE $W_{P'}\leftarrow{W_{P'}-w_i}$
		\ENDIF
		\STATE $P'\leftarrow{P'\setminus{\{i\}}}$, $n_{P'}\leftarrow{n_{P'}-1}$ \;
		\STATE Update $\theta$ and $\theta_v$ using Eqns. (\ref{eq:gradient-policy}) and (\ref{eq:gradient-value})
		\STATE $t\leftarrow{t+1}$
	\ENDWHILE
	\IF {$ov>Val[P']$} \STATE $Val[P']\leftarrow{ov}$
	\ENDIF
\ENDWHILE
\STATE \textbf{return} $Val$
\end{algorithmic}
\end{algorithm}

\subsection{State Aggregation}
\label{sec:state-aggregation}
As the number of items increases, the state space grows up exponentially and this affects the performance of function approximation with DNN. In order to shrink the state space and boost the method to have the capability of solving large problem instances, a new state embedding is derived by state aggregation. The feature values of states are divided into subsets and the values of each subset are converted to a certain value. Instead of manually testing different number of subsets to find the one that has the best performance, the process of finding appropriate number of subsets is considered as a sequential decision making problem and reinforcement learning is used for solving the problem. Before developing the RL framework, we first pre-process the problem instances. 

\paragraph{Preparing data.}
\label{sec:preparedata}
A set of problem instances are used for deriving the state embedding. Each problem instance is identified by a set of feature values which are the items information and capacity. The first step in aggregating the states is to generate random solutions for each problem instance. As mentioned before, an episode is a sequence of states and actions that each action selects an item and the solution is the set of selected items. These instances can be shown in a table in which each row corresponds to a problem instance and the columns are items information.

One issue in selecting the feature vector of original items information as states is that different KP instances are not comparable because the values and weights of items might be very different. As an example, assume that values and weights of an instance are integer numbers between 1 to 10, while these values and weights lies between 100 and 110 for another instance. Generalization based on these different values is difficult, although their ratio are similar. In order to solve this issue, for each item of instance $P$, all $v_i$ are normalized through dividing by the product of $w_i$ and $W_P$ as shown in Eqn. (\ref{eq:knapsacknormalized}). Furthermore, the ratio between $w_i$ and $W$ is also calculated based on Eqn. (\ref{eq:weightnormalized}). The $v_i$ and $w_i$ for each item are replaced with these two ratios in the feature vector of $P$. This modification makes the items of different problems comparable. The learned policy network in this way would boost the capability of the well-known heuristic greedy algorithm which is optimal for fractional KP.

\begin{equation}
\label{eq:knapsacknormalized}
vr_i(v_i, w_i, W_P) = \frac{v_i}{w_i \times{W_P}}
\end{equation}

\begin{equation}
\label{eq:weightnormalized}
wr_i(v_i, w_i, W_P) = \frac{w_i}{W_P}
\end{equation}

where, $vr_i$ and $wr_i$ are the normalized value and normalized weight respectively. For a problem instance $P$, $vr_i$, $wr_i$, $W_P$, $Sv$ and $Sw$ construct a feature vector $F(P)$.

\begin{equation}
\label{eq:featurevector}
F(P) = (F_1,...,F_{2N+4})=(n_P, W_P, Sv, Sw, vr_1, wr_1, ... , vr_{n_P}, wr_{n_P})
\end{equation}

where, $F(P)$ is the feature vector of $P$, $Sv$ and $Sw$ are the sum of remained values and weights respectively.

After obtaining a table of problem instances with comparable items, we sort for each row (i.e. each problem instance) the columns (i.e. $vr_i$ and $wr_i$) in descending order with respect to $vr_i$. In other words, the first two columns of each row, i.e. $vr_1$ and $wr_1$ correspond to the item with highest normalized value. The second two columns which are $vr_2$ and $wr_2$, correspond to the of item with the second highest normalized value and so on. For a problem $P$ with $n_P<N$, the items information are located from the columns $vr_1$ and $wr_1$ to $vr_{n_P}$ and $wr_{n_P}$ respectively. The values of $vr_{n_P+1}$ to $vr_N$ and also the values of $wr_{n_P+1}$ to $wr_N$ are zero. This ordering helps to aggregate all the highest $vr_i$ of all problem instances with a single aggregation strategy because the problem instances are comparable and the highest $vr_i$ is in a certain column. This explanation holds for second, third and other highest $vr_i$. Each column is called a feature and the next step is to derive an aggregation strategy for the values of each feature.

\paragraph{State aggregation through Q-Learning.}
\label{sec:rl-aggregator}

The idea of the aggregation is to reduce the number of unique values for all features. We do such reduction by splitting the values of one feature into several groups and then mapping each group's value to a particular integer. The proper number of splits for each feature is learned by reinforcement learning. For each feature $F_k$ that $k\in\{1,...,2N+4\}$, let action $d_{F_k}$ be the number of splits on the values of the feature $F_k$, and $F_{k,P}$ be the value of feature $F_k$ for problem instance $P$. Among all features, we perform state aggregation on $vr_i$ and $wr_i$ of item $i$. 

For aggregating the values of $vr_i$ of all $M$ problem instances, action $d_{vr_i}\in\{1,2,...,x\}$ is the number of splits where its optimal value i.e. $d^*_{vr_i}$ is obtained by Algorithm \ref{alg:aggregation}. Using $d^*_{vr_i}$ splits, the values of $vr_i$ are divided into $d^*_{vr_i}+1$ subsets and all the subsets except the last one have $\ceil[\big]{\frac{M}{d^*_{vr_i}+1}}$ values. The last subset has $M-(\ceil[\big]{\frac{M}{d^*_{vr_i}+1}}d^*_{vr_i})$ values. Then, all values of each subset is converted to an integer starting from 0. This process transforms the values of feature $vr_i$ to a set of integers $\{0,1,...,d^*_{vr_i}\}$ . As an example, assume there are $M=7$ problem instances that the values of $vr_1$ are $(1,2,6,3,1,2,5)$ and $d^*_{vr_1}$ is 2. These values need to be divided into $d^*_{vr_1}+1=3$ subsets. First they are sorted in order to acquire the sorted values $(1,1,2,2,3,5,6)$. Then, $3$ subsets $(\{1,1,2\},\{2,3,5\},\{6\})$ are obtained that each has $\ceil[\big]{7/3}=3$ values except the last one which has one value. Finally, the values of $vr_1$ are aggregated and the new values are $(0,0,2,1,1,0,1)$. The aggregation reduces  $5$ unique values of $vr_1$ to $3$ unique values.

For all $wr_i$, $d^*_{wr_i}$ is $2$ and the split points are 0.5 and 1. The motivation of this hard setting is separating illegal, light and heavy weights. Illegal weights are the weights with $wr_i>1$ that cannot be added to the knapsack. Similarly, $wr_i \leq 0.5$ and $0.5 < wr_i\leq 1$ determine light and heavy items respectively. The aggregation process is performed by the function $map(F_{k,P},d^*_{F_k})$ that gets $F_{k,P}$ and returns an integer which corresponds to a subset based on $d^*_{F_k}$ splits.

We used heuristics to define the reward function $R(F_k, d_{F_k})$ which is shown in Eqn. (\ref{eq:reward}).

\begin{equation}
\label{eq:reward}
R(F_k, d_{F_k}) = \frac{\prod_{j=1}^{d_{F_k}+1}{l_{F_k,j}}}{(d_{F_k}+1)\times{c_{F_k, d_{F_k}}}}
\end{equation}

where, $l_{F_k,j}$ is the size of $j^{th}$ subset, and $c_{F_k, d_{F_k}}$ is the number of all common values between all subsets. Three main motivations of designing rewards are as follows.

\begin{itemize}
\item We aim to define the reward function such that it reduces the size of state space. The number of unique states for each feature is $d_{F_k}+1$ after applying $d_{F_k}$ splits and this value inversely relates to the reward of each action.
\item For feature $F_k$ and $d_{F_k}$ splits, let $j\in\{0,1,2,...,d_{F_k}\}$ be a subset based on $d_{F_k}$ splits and $l_{F_k,j}$ be the difference between maximum and minimum values of $j^{th}$ subset. As larger $l_{F_k,j}$ entail in aggregating more values, their rewards are higher than those for smaller $l_{F_k,j}$. However, unequal subsets contain unequal number of values. For example, if the feature values are uniformly dispersed between 0 and 10, creating two subsets with lengths 5 and 5 are better than two subsets with length 1 and 9. Therefore, the product of the $l_{F_k,j}$ for all $j$ is in the numerator of the reward function.
\item Distinct states help an agent to derive a deterministic policy because states have dissimilar features. Likewise, two subsets with less overlapped values represent different sets of states and the policy can better distinguish them. For example, for the subsets $(\{1,1,2\},\{2,3,5\},\{6\})$, $2$ is common between two subsets and it can be assigned to both subsets. Assigning this value to different subsets entails a different policy that may have different performance. In order to reduce the number of common values between two groups, we define $c_{F_k, d_{F_k}}$ as the total number of common values in different subsets.
\end{itemize}
 
A $Q$ table is constructed for the states and actions and it is filled by the $Q-$learning algorithm \cite{sutton1998introduction} as shown in Algorithm \ref{alg:aggregation}. Each $vr_i$ is a state and the next state is the $vr_{i'}$ which $i'$ is an arbitrary state. Finally, an optimal decision is found by using the $Q$ table for each feature. The algorithm is used for aggregating $vr_i$ and we denote $d^*_{vr_i}$ as the optimal aggregation action for each $vr_i$. The state embedding derived by this strategy is a feature vector consisting of aggregated features and this state embedding is used in line \ref{line:s(p)} of algorithm \ref{alg:knapsack-solver}. Equation (\ref{eq:states}) shows $s(P)$, the state embedding of $P$.

\begin{equation}
\label{eq:states}
s(P) = \{map(F_{k,P}, d^*_{F_k}): \;\;\;\forall{F_k\in{F(P)}}\}
\end{equation}

\begin{algorithm}
\caption{Q-Learning for State Aggregation}
\label{alg:aggregation}
\textbf{Input}: Feature table of problem instances $P_1,...,P_M$\\
\textbf{Output}: The number of optimal split points for all $vr_i$ \\
\begin{algorithmic}[1] 
\STATE Initialize a $Q$ table with $N$ rows and $x$ columns. States are features and actions are the number of split points\;
\STATE Select item $i$ randomly \;

\REPEAT
	\STATE Select $i'$ randomly as the next item \;
	\STATE Select $d_{vr_i}\in{\{1,...,x\}}$ according to $\epsilon$-greedy policy \;
	\STATE Find $R(vr_i,d_{vr_i})$ using Eqn. \ref{eq:reward} \;
	\STATE Update $Q(vr_i,d_{vr_i})\leftarrow Q(vr_i,d_{vr_i})+\alpha [R_{vr_i,d_{vr_i}}+\gamma \max_{d'}{Q(vr_{i'},d')}-Q(vr_i,d_{vr_i})]$ \;
	\STATE $i = i'$ \;
\UNTIL{Convergence}
\STATE \textbf{return} $d^*_{vr_i}=argmax_d{Q(vr_i,d)}$  $\forall{i}$ 
\end{algorithmic}
\end{algorithm}

\section{Experiments}
\label{sec:experiments}
The DRL based knapsack solver is applied on three different types of problem instances. We tested different algorithms for training the policy DNN such as Deep Q Network (DQN) \cite{mnih2013playing}, Advantage Actor Critic (A2C) \cite{mnih2016asynchronous}, Proximal Policy Optimization \cite{schulman2017proximal} and Sample Efficient Actor-Critic with Experience Replay \cite{wang2016sample}, and the A2C algorithm is selected because it provided better solutions. We used \textit{stable-baseline} tools for implementing the A2C algorithm \cite{stable-baselines}. The policy network consists of two layers of 64 nodes and the method is trained on $10^4$ episodes which are selected from $M$ instances. The DRL with aggregation algorithm is compared with (1) greedy algorithm and (2) DRL without aggregation. The problem instances and code used for experiments are available in URL\footnote{The link is not shown due to the blind review.} .

\subsection{Problem Instances}
\label{sec:problem-instance}
The three different types of instances are called \textit{Random Instances}, \textit{Fixed $W_P$ Instances} and \textit{Hard Instances}. A set of $M$ problem instances makes a dataset that the maximum number of items over all instances in the dataset is $N$. Each dataset contains the instances of one of the following types.

\paragraph{Random instances (RI):}
A dataset of random instances has $M$ problem instances that each instance $P$ has $n_P\in{\{1,2,...,N\}}$ items. For an item $i$, $v_i$ and $w_i$ are randomly generated integers from one to $R$ that is a fixed upper bound for $v_i$ and $w_i$. The $W_P$ is a random integer between $R/10$ and $3R$. Three datasets of random instances are generated with $M=1000$. For these three datasets, $N$ is $50$, $300$ and $500$, and $R$ is $100$, $600$ and $1800$ respectively.

\paragraph{Fixed $W_P$ Instances (FI):}
In \cite{bello2017neural} a set of KP instances with fixed capacity and fixed item set size are used for evaluation. We generated three datasets of the same instances with $M=1000$. The $N$ for these three datasets is $50$, $300$ and $500$ respectively. The values and the weights of all items in the three datasets are random real numbers between zero and one. The $W_P$ is fixed for all the instances and it is 12.5 for $N=50$, 37.5 for $N=300$ and 37.5 for $N=500$.

\paragraph{Hard instances (HI):}
In \cite{pisinger2005hard}, a group of hard to solve problem instances were introduced that for each item $i$, $v_i$ is strongly correlated with $w_i$. Specifically, $w_i$ is a random integer in $[1,R]$, $v_i=w_i+R/10$ and $W_P=\frac{p}{M+1}\Sigma_{i=1}^{n_P}{w_i}$ where $p$ is the id of $P$. Three datasets of $M=1000$ hard instances are generated. For the first dataset, $N$ is 50 and $R$ is 100. Likewise, $N$ is 300 and 500, and $R$ is 600 and 1000 for the second and the third datasets respectively. 

\subsection{Evaluation Metrics}
\label{sec:metrics}
The following metrics are considered to evaluate the performance of using DRL based KP solver for solving the instances introduced in \ref{sec:problem-instance}.

\paragraph{Average values of solutions ($\overline{Val}$).}
For each dataset of $M$ problem instances, $\overline{Val}$ is the average of all solution values (total values of the selected items). 
Likewise, $\overline{Val}_{opt}$ is the average values of optimal solutions, which are obtained using the optimization solver \textit{Gurobi} \cite{gurobi}.

\paragraph{Learning rate. }
In order to calculate the learning rate, the rate of increasing in $Val$ is calculated per timesteps and the result is shown for each instance type when $N=300$.

\paragraph{Number of optimally solved instances ($\#_{opt}$).}
In order to evaluate the performance of the method on the individual problem instances, the number of instances that the method finds their optimal solution is computed for each dataset. 

\paragraph{Number of instances with highest solution value ($\#_{highest}$).}
This metric compares the value of solutions of DRL with aggregation and DRL without aggregation and counts the number of times that each one is higher. This value is calculated for the last $M/2$ instances of the datasets. These instances are more difficult to solve among the hard problem instances because  $W_P=\frac{p}{M+1}\Sigma_{i=1}^{n_P}{w_i}$ is increasing with respect to $p$.  Therefore, with larger $W_P$, the set of feasible solutions is bigger and hence finding the optimal solution is more difficult.

\begin{table*}[t]
\caption{Results of different algorithms and datasets of $M=1000$ problem instances. The method of \protect\cite{bello2017neural} is not applicable on RI and HI while it is optimal for small $N$ as well as the DRL w/ aggregation method. It is possible that two approaches find the optimal solution for a certain instance. Hence, the total number
 of optimally solved instances is not necessarily 1000.}
\label{tbl:statistics}
\centering
\begin{tabular}{llllllll}  
\hline
Dataset & Method & $N$ & $\overline{Val}$ & $\#_{opt}$ & $\#_{highest}$ & $\overline{Val}_{opt}$ & $\frac{\overline{Val}}{\overline{Val}_{opt}}$ \\
\hline
 & Greedy & 50 & 429.10 & 596 & 0 & 434.78 & 98.694\% \\
 & DRL w/o aggregation & 50 & 434.09 & 893 & 7 & 434.78 & 99.843\% \\
 & DRL w/ aggregation & 50 & \textbf{434.50} & \textbf{959} & \textbf{41} & 434.78 & 99.937\% \\
 & Greedy & 300 & 1144.96 & 418 & 0 & 1151.58 & 99.425\% \\
RI & DRL w/o aggregation & 300 & 1150.83 & 830 & 21 & 1151.58 & 99.934\% \\
 & DRL w/ aggregation & 300 & \textbf{1151.10} & \textbf{878} & \textbf{47} & 1151.58 & 99.958\% \\
 & Greedy & 500 & 15216.51 & 345 & 0 & 15285.56 & 99.548\% \\
 & DRL w/o aggregation & 500 & 15273.47 & 701 & 30 & 15285.56 & 99.920\% \\
 & DRL w/ aggregation & 500 & \textbf{15278.44} & \textbf{786} & \textbf{80} & 15285.56 & 99.953\% \\
\hline
 & Greedy & 50 & 20.10 & 172 & 0 & 20.15 & 99.738\% \\
 & DRL w/o aggregation & 50 & 20.14 & 740 & 36 & 20.15 & 99.931\% \\
 & DRL w/ aggregation & 50 & \textbf{20.15} & \textbf{773} & \textbf{54} & 20.15 & 99.959\% \\
 & Greedy & 300 & 86.26 & 202 & 0 & 86.31 & 99.942\% \\
 FI & DRL w/o aggregation & 300 & 86.27 & 226 & 24 & 86.31 & 99.961\% \\
 & DRL w/ aggregation & 300 & \textbf{86.29} & \textbf{330} & \textbf{205} & 86.31 & 99.976\% \\
 & Greedy & 500 & 111.68 & 204 & 0 & 111.73 & 99.945\% \\
 & DRL w/o aggregation & 500 & 111.63 & 64 & 31 & 111.73 & 99.871\% \\
 & DRL w/ aggregation & 500 & \textbf{111.70} & \textbf{261} & \textbf{144} & 111.73 & 99.970\% \\
\hline
 & Greedy & 50 & 772.428 & 134 & 0 &  802.72 & 96.226\% \\
 & DRL w/o aggregation & 50 & 799.036 & 655 & 113 & 802.72 & 99.540\% \\
 & DRL w/ aggregation & 50 & \textbf{799.438} & \textbf{689} & \textbf{147} & 802.72 & 99.591\% \\
 & Greedy & 300 & 27778.03 & 37 & 0 & 27965.76 & 99.328\% \\
HI & DRL w/o aggregation & 300 & 27947.11 & \textbf{370} & 161 &  27965.76 & 99.933\% \\
 & DRL w/ aggregation & 300 & \textbf{27952.63} & 336 & \textbf{233} & 27965.76 & 99.953\% \\
 & Greedy & 500 & 80779.23 & 25 & 0 & 81103.99 & 99.781\% \\
 & DRL w/o aggregation & 500 & 81022.60 & \textbf{217} & 168 & 81103.99 & 99.899\% \\
 & DRL w/ aggregation & 500 & \textbf{81064.99} & 166 & \textbf{304} & 81103.99 & 99.951\% \\
\hline
\end{tabular}
\end{table*}

\begin{figure}[h]
\centering
  \includegraphics[width=0.8\textwidth]{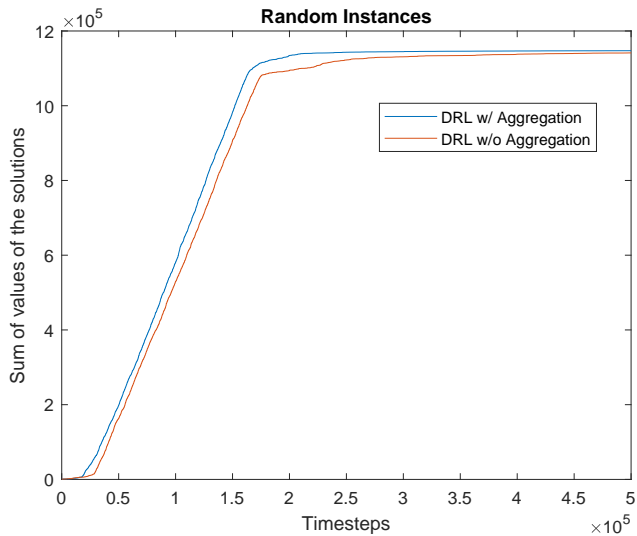}
\caption{Random Instances}
\label{fig:rndm}
\end{figure}
\begin{figure}
\centering
  \includegraphics[width=0.8\textwidth]{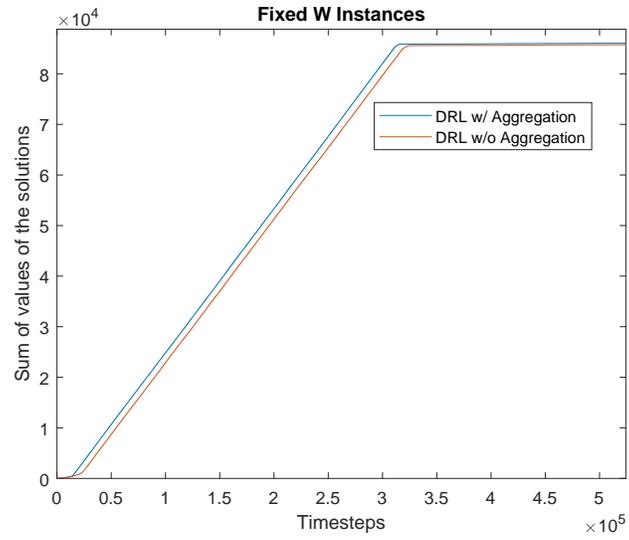}
\caption{Fixed $W_P$ instances}
 \label{fig:fixed}
  \end{figure}
  \begin{figure}
\centering
  \includegraphics[width=0.8\textwidth]{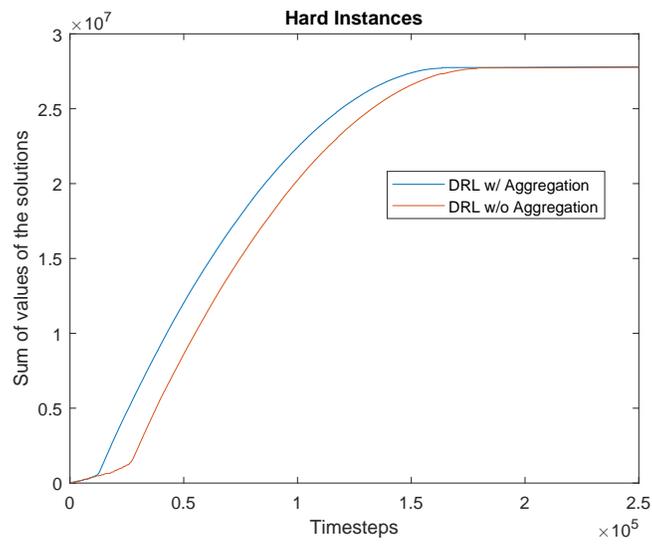}
\caption{Hard instances}
 \label{fig:hard}
\end{figure}

\subsection{Results} 

We ran the algorithm with 1000 problem instances. Table \ref{tbl:statistics} shows the quality of the solutions of different types of KP instances: RI, FI, and HI, that are obtained by DRL algorithms with (i.e. w/) and without (i.e. w/o) aggregation, and the greedy algorithm (Greedy).  

Table \ref{tbl:statistics} contains the ratio of $\overline{Val}$ and $\overline{Val}_{opt}$. These values show that the ratios of the solutions provided by our proposed method (DRL w/ aggregation) and the optimal solutions are most of the times more than $99.9\%$. This ratio does not change considerably when the number of items increases. 
Hence we conclude that our DRL based approach is able to find very close to optimal solutions for all instances we tested. 

\paragraph{Comparison with \cite{bello2017neural}.} The pointer network based DRL method \cite{bello2017neural} is also able to find close to optimal solutions for problem size up to $N=200$. 
However, the method of \cite{bello2017neural} can only be applied to solve the instances with exactly same number of items $N$, and in addition, with exactly same capacity value $W_P$.  In comparison, our DRL formulation allows to solve instances of any size up to and including $N=500$, and of any capacity value $W_P$. 

\paragraph{Comparison with Greedy and DRL without aggregation.}
The results show that the proposed DRL-based methods, with or without aggregation, always perform better than the greedy algorithm, in terms of the average solution quality ($\overline{Val}$), the number of optimally solved instances ($\#_{opt}$), and the number of instances with highest solution value ($\#_{highest}$).   

When we evaluate the advantage of having state aggregation, we notice that the state aggregation strategy improves the solutions especially for large instances, which is clearly observable in the solutions of instances RI and FI. 
Regarding to hard instances HI, the DRL with aggregation method 
is better than the one without aggregation strategy in terms of solution quality ($\overline{Val}$). For the large instances of sizes 300 and 500, the DRL without aggregation finds more optimal solutions than the one with aggregation, with 370 vs 336 for $N=300$, and 217 vs 166 for $N=500$. However, when looking at their performances in terms of how many times they have the highest solution values for 500 more difficult instances in HI, DRL with aggregation performs better than without aggregation, with 72 more wins for $N=300$, and 133 more wins for $N=500$. 
We investigate these instances in HI further. %
We have mentioned that 
$W_P$ is increasing with respect to $p$ which is the identifier of each problem instance $P$, for $M$ problem instances, this identifier ranges from one to $M$. When $p$ is small, $W_P$ is also small. Since $vr_i$ indirectly relates to $W_P$, it would be large when $W_P$ is small. For small $W_P$, the number of feasible solutions is low because less number of items can be fit into the knapsack. These problem instances are hence not actually ``hard''. In this case, 
state aggregation is not beneficial as  aggregating $vr_i$ of the items of the instances with small $p$ leads to sub-optimal solutions. 
However, problem instances with large $p$ 
have larger feasible solution space, and hence aggregation is beneficial as it reduces the state size to enhance generalization. 

The other benefit of DRL with aggregation method is that it is able to find the high quality solutions in less time steps. As it can be observed from Figures \ref{fig:rndm}, \ref{fig:fixed} and \ref{fig:hard}, the learning rate of DRL with aggregation method is higher than the DRL without aggregation. Hence, in general, it not only provides better solutions, but also the solutions are found in around $10,000$ less timesteps.

\section{Conclusion and future work}
\label{sec:conclusion}

In this paper we developed a DRL-based method for boosting the heuristic greedy algorithm and solving KP. In the DRL based KP solver, a policy DNN and a value DNN are trained using A2C algorithm and the policy DNN is used for sequentially selecting items to find a solution. The states in DRL modeling of KP contain the information of the instances that are aggregated to reduce the state space. The state aggregation policy is derived by solving a tabular RL problem. Using this aggregation policy, a state embedding is obtained and this state embedding is used with another RL framework to train the parameters of the policy network.

We applied this method on three types of problem instances named random instances, fixed $W_P$ instances and hard instances. Three datasets with 50, 300 and 500 are generated for each type. The DRL with aggregation method found close to optimal solutions for the instances. It also found optimal solutions for fixed capacity instance with small number of items as well as the method developed by \cite{bello2017neural}.

The proposed method can be generalized to other COPs. For instance, the TSP consists of some cities and the goal is to find the minimum length tour that visits every city exactly once. The cities might be the items and an aggregation strategy could reduce the state space by aggregating the coordinates. As another example, in minimum vertex cover problem, the items may be the vertices and aggregation can be performed by grouping the weight of the vertices. 

In this paper, we use RL to automate the reduction of the state space, as a pre-processing step of the DRL based approach for KPs. It is also interesting to investigate in the future how better reward functions can be derived through learning. This might be very helpful for problems that many tuning processes are needed for deriving a strong reward function. In general, automating the state, reward and action derivation for RL problems are interesting topics for research in the future.



%
%
%
\bibliographystyle{splncs04}
\bibliography{ecml20}
\end{document}